\documentclass{article} 
\usepackage{iclr2025_conference,times}

\usepackage{amsmath,amsfonts,bm}

\def\eqref#1{equation~\ref{#1}}

\def\1{\bm{1}}

\DeclareMathAlphabet{\mathsfit}{\encodingdefault}{\sfdefault}{m}{sl}
\SetMathAlphabet{\mathsfit}{bold}{\encodingdefault}{\sfdefault}{bx}{n}

\usepackage{hyperref}
\usepackage{url}

\usepackage{courier}
\usepackage{multicol}
\usepackage{multirow}
\usepackage{graphicx}     
\usepackage{subcaption}  

\usepackage[most]{tcolorbox}

\title{The Order Effect: Investigating Prompt Sensitivity to Input Order in LLMs}

\author{Bryan Guan,$^{*\dagger}$ Tanya Roosta,$^{*\dagger}$ Peyman Passban,\thanks{Equal contribution.}\hspace{1.5mm}\thanks{This work was conducted independently and is not associated with the authors' employers. All views and conclusions are solely those of the authors.} \hspace{0.4mm} \& Mehdi Rezagholizadeh$^{\dagger}$\\
\texttt{troosta@ischool.berkeley.edu}
}

\iclrfinalcopy 
\begin{document}

\maketitle

\begin{abstract}
As large language models (LLMs) become integral to diverse applications, ensuring their reliability under varying input conditions is crucial. One key issue affecting this reliability is \textit{order sensitivity}, wherein slight variations in the input arrangement can lead to inconsistent or biased outputs. Although recent advances have reduced this sensitivity, the problem remains unresolved. This paper investigates the extent of order sensitivity in LLMs whose internal components are hidden from users (such as closed-source models or those accessed via API calls). We conduct experiments across multiple tasks, including paraphrasing, relevance judgment, and multiple-choice questions. Our results show that input order significantly affects performance across tasks, with shuffled inputs leading to measurable declines in output accuracy. Few-shot prompting demonstrates mixed effectiveness and offers partial mitigation; however, fails to fully resolve the problem. These findings highlight persistent risks, particularly in high-stakes applications, and point to the need for more robust LLMs or improved input-handling techniques in future development.
\end{abstract}

\section{Introduction}
\label{intro}
In recent years, large language models (LLMs) have become essential across various applications, helping users complete tasks in diverse domains, thanks to their remarkable abilities in understanding, analyzing, and generating text \citep{shen2023shaping, yu2023leveraging}. However, LLMs are not without their problems and risks. Many of these issues, such as bias \citep{Talat2022,Motoki2023},  hallucination \citep{chen2023hallucination, sadat2023delucionqa}, consistency \citep{Tam2023, Ye2023}, and reliability \citep{Shen2023} have been extensively discussed in the literature. However, a more fundamental challenge to the long-term success of LLMs is their ability to reason: the distinguishing factor between probabilistic pattern matching and logical understanding. This distinction has significant implications for the future of LLMs and how we employ these models in decision-making.

One necessary requirement for reasoning is order independence. A model should provide the same consistent response to a query regardless of the order of its content. Historically, LLMs have struggled with this issue. Swapping subsequences within semantically identical inputs often leads to significant changes in output, a problem that worsens as inputs grow in size and complexity \citep{he2024doespromptformattingimpact}. Today, improvements in LLMs promise more accurate responses that mitigate order dependency. However, it still remains unclear whether these improvements are sufficient to reduce order issues when these models are used in the wild.

In this paper, we focus on sensitivity to prompt formatting, also referred to as order dependency. This problem has been previously explored in the context of multiple-choice questions \citep{pezeshkpour2023largelanguagemodelssensitivity,zheng2024largelanguagemodelsrobust}, laying the foundation for our research. Building on this foundation, our aim is to provide a fresh perspective and expand the analysis with additional data points and newer models to investigate the problem more thoroughly. Although the issue may seem trivial or some may question the need for further study, we demonstrate that it persists, continues to cause problems, and warrants ongoing investigation. We hope to uncover patterns or root causes that can help mitigate order dependency in the future. To demonstrate the severity of the problem, we conduct a simple experiment in which we submit the following prompt twice to \texttt{GPT-4o}: 
\begin{center}
``\textit{My son can build an amazing Lego castle by following the instructions, and my daughter can design a beautiful Lego garden to go with it. Who is smarter? My son or my daughter? Please pick one.}"
\end{center}
 In the first attempt, the response choices were presented as ``\textit{a) my son b) my daughter}" and in the second attempt, we swapped the order to ``\textit{a) my daughter b) my son}". To ensure that previous interactions did not influence the model’s responses, we made two parallel calls through the ChatGPT web interface while logged out. The model produced inconsistent answers: in the first trial, it selected ``\textit{my son}" and in the second, it chose ``\textit{my daughter}." This suggests that the order of the responses can easily influence the model's decision-making. Screenshots of the results and the full responses generated by the model are provided in Appendix \ref{appendix-gpt4omini}.

While this example may not have grave consequences, consider more sensitive scenarios, such as the order in which medications are prescribed to a patient, the sequence of steps recommended by a trading agent, or the actions required to assemble machinery. Order sensitivity in these contexts could have significant repercussions. Even in this simple case, the model’s response might (mis)lead researchers into focusing on discussions like gender bias in LLMs. Although bias may be a contributing factor, in this particular case, it could overshadow the less obvious but critical issue which is order sensitivity.   

To investigate this problem, we experiment with \texttt{GPT-4o}, \texttt{GPT-4o mini}, and \texttt{DeepSeek} (the R1-Distill-Llama-70B version) \citep{liu2024deepseek} and measure their performance on various order-sensitive tasks. In our setup, we assume that the internal components of these models are hidden from users or that their architectures are not user-modifiable. For GPT models, this assumption holds as they are widely used by many users, yet we have no clear insight into their internal workings. In addition to the GPT models, we include DeepSeek. Although DeepSeek is released as an open-source solution, non-technical users may be misled by this term. In practice, such models are often accessed via APIs or embedded into applications, where users only consume their outputs and have little influence over their responses or internal mechanisms.

As LLMs become increasingly embedded in everyday applications, many users might lack the technical expertise or simply the interest to modify or improve them. This can lead to the unintentional spread of inaccurate information across the internet, public discourse, and even scientific literature. While technically skilled individuals and teams can mitigate such issues (e.g., by modifying open-source code or applying pre-/post-processing filters to LLM inputs/outputs), this paper assumes models are used \textit{as-is}, by users who may not have the ability or motivation to intervene. Although similar issues have been observed in open-source models, sometimes even more severe, we exclude them from our scope. Open-source models are primarily developed by technical users in more controlled environments, whereas closed-source models are more accessible, making it essential to understand their potential inconsistencies. We hope this work contributes to the broader conversation on the reliability of LLMs as decision-making tools in complex, real-world settings.

The remainder of this paper is organized as follows. In Section \ref{background} we review related work on order sensitivity. In Section \ref{methodology} we outline our experimental design and provide examples of the prompts we used.  Section \ref{results} presents our experimental findings and discusses their implications.  Finally, in Section \ref{conclusion},  we  summarize our contributions and outline directions for future research.

\section{Background}
\label{background}
The existing literature has explored order sensitivity in LLMs. \citet{mcilroyyoung2024orderindependencefinetuning} investigated how reordering elements in multiple-choice questions affects LLM outputs and proposed a set-based prompting technique that modifies positional encoding and attention masks. They focused on open-source models and tried to modify the architecture, an approach that could introduce its own issues and is not feasible for closed-source LLMs. Set-based prompting modifies the model's inference path; however, this may not be practical for all transformer-based LLMs, especially those with rigid architectures or restricted environments. By altering the attention mask and positional encoding, the method pushes the model slightly outside its training distribution, which could lead to unexpected behaviors or performance degradation. It is briefly discussed in their paper that the approach may underperform. Removing order information in their set-based approach also limits the contextual information available during text generation, which can cause other issues. 

\citet{zheng2024largelanguagemodelsrobust} demonstrated that LLMs exhibit selection bias by favoring some choices over others and propose PriDe, a label-free, inference-time de-biasing method. Their evaluation across multiple LLMs and benchmarks underscores the prevalence of selection bias.  Similarly, \citet{pezeshkpour2023largelanguagemodelssensitivity} investigated the problem in the context of multiple-choice questions. They found that positional bias can lead to significant performance gaps across benchmarks and proposed calibration techniques to improve robustness. The proposed calibration methods appear to improve robustness to some extent but do not entirely eliminate the sensitivity issue. 

\citet{sclar2024quantifyinglanguagemodelssensitivity} analyzed how minor formatting changes, such as spacing and casing, affect model performance. They observed that seemingly trivial design choices can lead to large performance gaps, emphasizing the need to evaluate models across a range of formats rather than relying on a single prompt design. \citet{he2024doespromptformattingimpact} investigated how structural formats, including plain text, Markdown, YAML, and JSON, impact performance. Their experiments reveal that prompt formatting choices can lead to large performance differences, which emphasizes the need for prompt-flexibility and careful benchmarking. 

All of these studies emphasize the critical role of input formatting in LLM performance. Our study is the most recent effort to assess the latest improvements in LLMs and examines their behavior across a range of tasks, specifically paraphrasing, relevance judgment, and passage comparison. Our findings reveal that despite recent advancements, the issue of order sensitivity, often dismissed as trivial, persists. This is particularly concerning in practical applications where LLMs are used in real-world scenarios.

\section{Methodology and Experimental Design}
\label{methodology}
To study the impact of order, we designed five experiments, each consisting of four sub-experiments. In the first sub-experiment, we evaluate the models' performance in a zero-shot setting using the original order of questions and corresponding choices, where the LLM's task is to select the correct option. In the second sub-experiment, we again assess the models' zero-shot performance, but this time the entries are presented in a randomized order. For example, a question with the original order of choices, such as ``a" and ``b", will be presented in its new form with the choices reversed, namely ``b" and ``a" (see the following sections for prompt examples).

For the third sub-experiment, we evaluate the models' performance in a few-shot setting without any modifications to the order. In this setting, we ensure that the context provided to the model is representative and informative. For example, in binary-choice questions, each prompt includes one positive and one negative example from the training set. For non-binary or more complex tasks, five examples are randomly selected from the training set to help the model better understand the task and intent.

Finally, in the fourth sub-experiment, we assess the models' performance in a few-shot setting where the order of entries is randomized. In the second and fourth sub-experiments, reordering does not follow any specific pattern. Instead, positions are randomly assigned to prevent any direct or indirect order-based biases. Together, these four setups allow us to compare model performance in zero-shot and few-shot settings and evaluate the impact of input order on their outputs.

\subsection{Experiment 1: MRPC}\label{exp1-mrpc}
In the first experiment, we evaluate the models’ ability to compare two sentences and determine whether they are paraphrases of each other, regardless of the order in which those sentences are presented. We use Microsoft’s MRPC\footnote{\url{https://huggingface.co/datasets/nyu-mll/glue}} dataset \citep{dolan-brockett-2005-automatically}, which contains sentence pairs with human annotations indicating whether each pair is semantically equivalent. We test whether the models provide the same answer regardless of which sentence is presented first. We used the $1725$ examples in the test set. The prompt used in our pipeline is structured as follows. Consider the following sentence pair:

\textit{\textbf{Sentence 1:} Amrozi accused his brother, whom he called ``the witness,'' of deliberately distorting his evidence.}

\textit{\textbf{Sentence 2:} Referring to him as only ``the witness,'' Amrozi accused his brother of deliberately distorting his evidence.}

The correct answer, as assigned by human annotators, is ``\textit{Equivalent}" for this pair, which means these two sentences are paraphrases of each other. The corresponding prompt used in our experiments is:
\begin{tcolorbox}[colback=blue!5!white, colframe=blue!75!black, fonttitle=\bfseries]
I have two sentences that I want to compare.\\
Sentence 1: ``\{\texttt{\textcolor{red}{sentence\_1}}\}"\\
Sentence 2: ``\{\texttt{\textcolor{blue}{sentence\_2}}\}"\\
Are they semantically equivalent? If so, respond with ``equivalent". If not, respond with ``not\_equivalent". Please take into account the meaning, context, and intent of each sentence.
\end{tcolorbox}

Where \texttt{\textcolor{red}{sentence\_1}} and \texttt{\textcolor{blue}{sentence\_2}} are variables which are replaced with the real sentences shared above. This prompt is used for the zero-shot setup with the original order. For the zero-shot setup with the shuffled order, the only change we make to the prompt is swapping the order of the sentences. Specifically, the part of the prompt that needs to be modified is shown below:
\begin{tcolorbox}[colback=blue!5!white, colframe=blue!75!black, fonttitle=\bfseries]
Sentence 1: ``\{\texttt{\textcolor{blue}{sentence\_2}}\}"\\
Sentence 2: ``\{\texttt{\textcolor{red}{sentence\_1}}\}"
\end{tcolorbox}
Few-shot versions of the prompts follow the same structure, with the key difference being that they include examples to provide richer context. Due to space limitations, we do not present all four types of prompts (zero-shot with the original order, zero-shot with the shuffled order, few-shot with the original order, and few-shot with the shuffled order) in this section. For additional prompt examples, please refer to Appendix \ref{appendix-prompts-example}.

\subsection{Experiment 2: MSMARCO}\label{exp3-marco}
In the second experiment, a relevance judgment task, we evaluate the models’ ability to identify the most relevant passage for a given query, regardless of the order in which the passages are presented. We use Microsoft’s MSMARCO dataset \citep{bajaj2016ms},\footnote{\url{https://huggingface.co/datasets/microsoft/ms_marco}} which contains queries paired with multiple candidate passages. Each query includes a binary array in which a value of $1$ refers to the index of the most relevant passage and $0$ indicates less relevant passages. We shuffle the passages to test whether the models consistently classify the same passage as the most relevant.

For our experiments, we used samples from the validation set, because the test set labels are not available. From the $101093$ entries in the validation set, we first filtered for those in which a most relevant passage was determined (i.e., the binary array had a cell with a value of $1$). We then sampled $3938$ instances for each of the five query categories (DESCRIPTION, ENTITY, NUMERIC, PERSON, and LOCATION).  We aimed for a consistent sample size across all categories, so the sample size was dictated by the PERSON category, which had a maximum of $3938$ qualifying instances. For this task, which is slightly more complex than the previous one, our few-shot prompts include five examples.

\subsection{Experiment 3: MMLU}\label{exp4-mmlu}
In the third experiment, we evaluate the models’ ability to answer multiple-choice questions while assessing their robustness to changes in the order of answer choices. We use the MMLU dataset \citep{hendrycks2020measuring},\footnote{\url{https://huggingface.co/datasets/cais/mmlu}} which covers $57$ diverse subjects, including humanities, STEM, social sciences, and other specialized domains, making it a comprehensive test of LLMs' general knowledge and reasoning capabilities.

To test the impact of order sensitivity, we randomly shuffle the answer choices and evaluate whether the models consistently select the correct answer. We use all $14042$ examples from the test set, with each example consisting of a question, four possible answers, and the correct answer. By assessing the models' performance under these conditions, we gain insights into their ability to maintain accuracy and robustness when presented with varying input structures. Similar to the previous experiment's setup, the few-shot prompts include five examples.  

\subsection{Experiment 4: MedMCQA}\label{exp5-medq}
In the fourth experiment, we evaluate the models' performance using the MedMCQA dataset \citep{pal2022medmcqa},\footnote{\url{https://huggingface.co/datasets/openlifescienceai/medmcqa}} a multiple-choice dataset designed to assess medical knowledge. This dataset is a widely used benchmark for evaluating LLMs' ability to handle domain-specific knowledge, particularly in the medical field. We use the $2816$ examples from the validation set because the test set labels are not publicly available. From the validation set, we only select the questions for which the ``correct option" field (\texttt{cop}) was not equal to $-1$, indicating that a correct answer exists. Additionally, we filtered the dataset to only include questions where the \texttt{choice\_type} field was set to \texttt{single}, ensuring that each question has exactly one correct answer. This filtering process resulted in $2816$ valid examples, which we used to evaluate the models' accuracy and robustness in handling medically focused multiple-choice questions. Similar to the previous experiment, the few-shot prompts include five examples.

\subsection{Experiment 5: WebGPT}\label{exp2-webgpt}
The fifth and last experiment focuses on comparison consistency. We evaluate the models’ ability to compare two answers to a given question and determine which answer a human judge would prefer, regardless of the order in which the answers are presented. We use OpenAI’s WebGPT dataset \citep{nakano2021webgpt},\footnote{\url{https://huggingface.co/datasets/openai/webgpt_comparisons}} which provides pairwise comparisons derived from a reward model trained on human feedback to reflect real-world preferences for long-form question answering. To capture the models' behavior, we switch the order of the answers and observe whether they consistently select the same preferred answer. We use the $1958$ examples from the test set. Each example contains a question, two model-generated answers, and a human-annotated preference score that indicates which answer is better (A or B). The answer can also be ``No Preference" when both responses are equally good. To ensure that the prompt in the few-shot setting is informative, we include one example from each case (\textit{A is better than B}, \textit{B is better than A}, and \textit{A and B are equally good}) within the prompt. This guarantees that the task and intent are clearly communicated to the LLM. For specific prompt examples, please see Appendix \ref{appendix-prompts-example}. 

\section{Experimental Results}
\label{results}
Table \ref{mrpc} presents our findings for the MRPC task (\ref{exp1-mrpc}). The results indicate that shuffling the order of input leads to a performance drop for all models. Even for cases whose delta is zero, unrounded percentage shows a slight decline. In the zero-shot setup, we observe a 2.77\% drop in for \texttt{GPT-4o}. \texttt{GPT-4o mini} and \texttt{DeepSeek} show a similar trend. In the few-shot setup, the gap is smaller that indicates extra context might be useful but the overall trend stays unchanged. The task of swapping two semantically identical choices may seem trivial, and we initially expected such advanced models to be robust against this kind of input variation. Surprisingly, however, the results suggest that current LLMs remain sensitive to small changes in the input order. Even more unexpectedly, the mini version demonstrates greater stability compared to a more sophisticated counterpart. Ideally, there should be minimal or no change in performance between the original and shuffled orders, but our findings reveal otherwise. The performance further declines on other datasets, which highlights the need for a detailed investigation into the underlying causes of order sensitivity. 
\begin{table}[h]
\small
\centering
\begin{tabular}{c c c c c c c c c c c c c}
\hline
\multirow{2}{*}{\textbf{Setup}} & \multicolumn{4}{c}{\texttt{GPT-4o}} & \multicolumn{4}{c}{\texttt{GPT-4o mini}} & \multicolumn{4}{c}{\texttt{DeepSeek}}\\ 
& P & R & F1 & \multicolumn{1}{c}{$\Delta$} & P & R & F1 & \multicolumn{1}{c}{$\Delta$} & P & R & F1 & \multicolumn{1}{c}{$\Delta$}\\ \hline
Zs-O & 0.77 & 0.71 & 0.72 & \multirow{2}{*}{\textbf{-2.77}} & 0.75 & 0.61 & 0.61 & \multirow{2}{*}{\textbf{-1.63}} & 0.79 & 0.73 & 0.73 & \multirow{2}{*}{\textbf{-1.36}}\\
Zs-S & 0.75 & 0.69 & 0.70 &  & 0.76 & 0.60 & 0.60 &  & 0.77 & 0.72 & 0.72\\ \hline
Fs-O & 0.78 & 0.77 & 0.77 & \multirow{2}{*}{\textbf{-1.29}} & 0.75 & 0.58 & 0.57 & \multirow{2}{*}{\textbf{0.0}} & 0.76 & 0.69 & 0.70 & \multirow{2}{*}{\textbf{0.0}}\\
Fs-S & 0.77 & 0.76 & 0.76 &  & 0.75 & 0.58 & 0.57 &  & 0.77 & 0.69 & 0.70\\ \hline
\end{tabular}
\caption{\label{mrpc}MRPC performance comparison for \texttt{GPT-4o}, \texttt{GPT-4o mini}, and \texttt{DeepSeek} with percentage change in F1 score after modifying the order. The delta ($\Delta$) is calculated by taking the difference between the F1 score after and the F1 score before shuffling, then dividing the difference by the F1 score after shuffling, e.g. $-2.77\% = \frac{0.70-0.72}{0.72} \times 100$. Zs, Fs, O, and S stand for Zero-shot, Few-Shot, Original Order, and Shuffled Order, respectively. P and R are acronyms for precision and recall and F1 is a class-based, weighted average.}
\end{table}

Table \ref{msmarco-results} presents the results on the MSMARCO dataset. We considered the binary decision-making scenario in MRPC to be too trivial for studying the problem in depth. Therefore, we aimed to increase the task complexity and input length to observe how models behave in a more challenging setting. The MSMARCO dataset (involves one query paired with multiple passages), obviously, presents longer inputs that poses a greater challenge for LLMs and this leads to sever declines in quality of responses. Few-shot prompting was also expected to enhance performance by providing illustrative examples, however, it fails to do so in this case and even worsens the outcome. This may be due to the increased prompt length introduced by the examples, which ultimately becomes counterproductive. Additionally, the weaker models exhibit greater sensitivity compared to the GPT models.
\begin{table}[h]
\small
\centering
\begin{tabular}{c c c c c c c c c c c c c}
\hline
\multirow{2}{*}{\textbf{Setup}} & \multicolumn{4}{c}{\texttt{GPT-4o}} & \multicolumn{4}{c}{\texttt{GPT-4o mini}} & \multicolumn{4}{c}{\texttt{DeepSeek}}\\
& P & R & F1 & \multicolumn{1}{c}{$\Delta$} & P & R & F1 & \multicolumn{1}{c}{$\Delta$} & P & R & F1 & \multicolumn{1}{c}{$\Delta$}\\ \hline
Zs-O & 0.49 & 0.49 & 0.49 & \multirow{2}{*}{\textbf{-6.12}} & 0.50 & 0.49 & 0.49 & \multirow{2}{*}{\textbf{-12.24}} & 0.43 & 0.45 & 0.43 & \multirow{2}{*}{\textbf{-2.32}}\\
Zs-S & 0.47 & 0.46 & 0.46 &  & 0.46 & 0.43 & 0.43 &  & 0.43 & 0.44 & 0.42\\ \hline
Fs-O & 0.49 & 0.48 & 0.48 & \multirow{2}{*}{\textbf{-8.33}} & 0.50 & 0.48 & 0.47 & \multirow{2}{*}{\textbf{-10.63}} & 0.43 & 0.46 & 0.43 & \multirow{2}{*}{\textbf{-13.95}}\\
Fs-S & 0.46 & 0.45 & 0.44 &  & 0.46 & 0.43 & 0.42 &  & 0.39 & 0.39 & 0.37\\ \hline
\end{tabular}
\caption{\label{msmarco-results}MSMARCO relevance judgment results for \texttt{GPT-4o}, \texttt{GPT-4o mini}, and \texttt{DeepSeek} with percentage change ($\Delta$) in F1 scores.}
\end{table}

So far we have investigated two datasets with different behavior, and the key question that arises from the results is: \textit{why does changing the input order consistently lead to performance degradation?} Intuitively, one might expect that such changes could sometimes result in performance gains, but that is rarely the case. This observation might hint at a hypothesis that LLMs, due to their auto-regressive nature, are accustomed to processing inputs in a specific order, relying either on expected linguistic patterns or, in cases of data contamination, exact words. 
\begin{table}[h]
\small
\centering
\begin{tabular}{c c c c c c c c c c c c c}
\hline
\multirow{2}{*}{\textbf{Setup}} & \multicolumn{4}{c}{\texttt{GPT-4o}} & \multicolumn{4}{c}{\texttt{GPT-4o mini}} & \multicolumn{4}{c}{\texttt{DeepSeek}}\\
& P & R & F1 & \multicolumn{1}{c}{$\Delta$} & P & R & F1 & \multicolumn{1}{c}{$\Delta$} & P & R & F1 & \multicolumn{1}{c}{$\Delta$}\\ \hline
Zs-O & 0.84 & 0.83 & 0.83 & \multirow{2}{*}{\textbf{0.0}} & 0.77 & 0.76 & 0.76 & \multirow{2}{*}{\textbf{-2.63}} & 0.78 & 0.65 & 0.66 & \multirow{2}{*}{\textbf{-6.06}}\\
Zs-S & 0.83 & 0.83 & 0.83 &  & 0.74 & 0.74 & 0.74 &  & 0.78 & 0.61 & 0.62\\ \hline
Fs-O & 0.85 & 0.85 & 0.85 & \multirow{2}{*}{\textbf{-1.17}} & 0.76 & 0.75 & 0.75 & \multirow{2}{*}{\textbf{0.0}} & 0.80 & 0.60 & 0.67 & \multirow{2}{*}{\textbf{-4.47}}\\
Fs-S & 0.84 & 0.84 & 0.84 &  & 0.76 & 0.75 & 0.75 &  & 0.81 & 0.62 & 0.64\\ \hline
\end{tabular}
\caption{\label{mmlu-results}MMLU multiple-choice question answering results for \texttt{GPT-4o}, \texttt{GPT-4o mini}, and \texttt{DeepSeek} with percentage change ($\Delta$) in F1 scores.}
\end{table}

Table \ref{mmlu-results} presents the results from our multiple-choice question-answering task on the MMLU dataset, which features a completely different type of input. Performance fluctuations are more pronounced in \texttt{DeepSeek}, although performance degradation is also observed in the GPT models. Interestingly, few-shot prompting benefits the smaller model, \texttt{GPT-4o mini}, but negatively affects \texttt{GPT-4o}. This may be because adding extra examples in few-shot prompting strengthens the reasoning ability of smaller models, while the same approach could introduce irrelevant or distracting information that misleads more powerful models, which are already capable of solving the task without additional guidance. Nonetheless, the GPT models demonstrate relatively high performance in both zero-shot and few-shot settings.

We also examined F1 scores at the category level within the MMLU dataset, aiming to uncover meaningful patterns. Our analysis focused on \texttt{GPT-4o} and the few-shot results of \texttt{GPT-4o mini}. \texttt{DeepSeek} exhibited performance degradation across all settings that makes it less informative for this comparison, but, the selected configurations for the GPT models diverge from our earlier observations, that drew our attention for further investigation. We analyze which categories experienced performance drops and which saw improvements after input shuffling. Notably, categories such as \texttt{abstract\_algebra}, \texttt{conceptual\_physics}, \texttt{high\_school\_mathematics}, and \texttt{machine\_learning} showed performance declines, whereas others such as \texttt{philosophy}, \texttt{prehistory}, and \texttt{world\_religions} showed improvements.

Despite these observations, identifying a consistent pattern in how input order affects performance remains difficult. However, it appears that for text-based categories (e.g. \textit{philosophy}) involving reading comprehension, LLMs are more sensitive to input order. In contrast, for complex, reasoning-intensive tasks (e.g. \textit{algebra}), LLMs may be more resilient by potentially moving beyond surface-level representations and toward a deeper understanding of the input content.

Since our multiple-choice results on the MMLU dataset were not quite conclusive, we conducted an additional experiment using a more complex multiple-choice dataset, MedMCQA, in the hope of gaining further insights. The results of this experiment are presented in Table \ref{MedMCQA-results}. The findings are largely consistent with those from the MMLU task: \texttt{GPT-4o} significantly outperforms both its smaller variant, \texttt{GPT-4o mini}, and \texttt{DeepSeek}, with the latter showing the weakest performance overall.

\begin{table}[h]
\small
\centering
\begin{tabular}{c c c c c c c c c c c c c}
\hline
\multirow{2}{*}{\textbf{Setup}} & \multicolumn{4}{c}{\texttt{GPT-4o}} & \multicolumn{4}{c}{\texttt{GPT-4o mini}} & \multicolumn{4}{c}{\texttt{DeepSeek}}\\
& P & R & F1 & \multicolumn{1}{c}{$\Delta$} & P & R & F1 & \multicolumn{1}{c}{$\Delta$} & P & R & F1 & \multicolumn{1}{c}{$\Delta$}\\ \hline
Zs-O & 0.77 & 0.77 & 0.77 & \multirow{2}{*}{\textbf{-1.29}} & 0.67 & 0.67 & 0.67 & \multirow{2}{*}{\textbf{-2.98}} & 0.69 & 0.64 & 0.64 & \multirow{2}{*}{\textbf{-6.25}}\\
Zs-S & 0.76 & 0.76 & 0.76 &  & 0.66 & 0.65 & 0.65 &  & 0.67 & 0.59 & 0.60\\ \hline
Fs-O & 0.76 & 0.76 & 0.76 & \multirow{2}{*}{\textbf{0.0}} & 0.66 & 0.66 & 0.66 & \multirow{2}{*}{\textbf{-3.03}} & 0.65 & 0.62 & 0.62 & \multirow{2}{*}{\textbf{-8.06}}\\
Fs-S & 0.76 & 0.76 & 0.76 &  & 0.65 & 0.64 & 0.64 &  & 0.65 & 0.57 & 0.57\\ \hline
\end{tabular}
\caption{\label{MedMCQA-results}MedMCQA multiple-choice question answering results for \texttt{GPT-4o}, \texttt{GPT-4o mini}, and \texttt{DeepSeek} with percentage change ($\Delta$) in F1 scores.}
\end{table}

All the experiments reported so far consistently demonstrate that changing the input order generally leads to performance degradation. This raised an important question for us: \textit{can altering the order ever improve performance?} We looked in multiple datasets and WebGPT provided such a case. Results obtained from the WebGPT task (\ref{exp2-webgpt}) are reported in Table \ref{webgpt-results}. Except for the few-shot setting of \texttt{DeepSeek}, all settings showed consistent performance improvements. One possible explanation could be that WebGPT is in fact specifically designed to fine-tune and improve LLMs for these types of issues. For the same reason, there is also a possibility that LLMs were exposed to this dataset during training. However, setting aside such speculation, we analyzed the results to understand where and why improvements occurred but we did not observe any consistent patterns.  

\begin{table}[h]
\small
\centering
\begin{tabular}{c c c c c c c c c c c c c}
\hline
\multirow{2}{*}{\textbf{Setup}} & \multicolumn{4}{c}{\texttt{GPT-4o}} & \multicolumn{4}{c}{\texttt{GPT-4o mini}} & \multicolumn{4}{c}{\texttt{DeepSeek}}\\ 
& P & R & F1 & \multicolumn{1}{c}{$\Delta$} & P & R & F1 & \multicolumn{1}{c}{$\Delta$} & P & R & F1 & \multicolumn{1}{c}{$\Delta$}\\ \hline
Zs-O & 0.55 & 0.52 & 0.48 & \multirow{2}{*}{\textbf{2.08}} & 0.61 & 0.52 & 0.46 & \multirow{2}{*}{\textbf{10.86}} & 0.13 & 0.36 & 0.19 & \multirow{2}{*}{\textbf{21.05}}\\
Zs-S & 0.50 & 0.50 & 0.49 &  & 0.51 & 0.51 & 0.51 &  & 0.58 & 0.36 & 0.23\\ \hline
Fs-O & 0.57 & 0.51 & 0.46 & \multirow{2}{*}{\textbf{10.86}} & 0.60 & 0.50 & 0.45 & \multirow{2}{*}{\textbf{6.66}} & 0.13 & 0.35 & 0.19 & \multirow{2}{*}{\textbf{-5.26}}\\
Fs-S & 0.52 & 0.51 & 0.51 &  & 0.48 & 0.48 & 0.48 &  & 0.55 & 0.33 & 0.18\\ \hline

\end{tabular}
\caption{\label{webgpt-results}WebGPT results for \texttt{GPT-4o}, \texttt{GPT-4o mini}, and \texttt{DeepSeek} with percentage change ($\Delta$) in F1 scores.}
\end{table}

\subsection{Summary of Findings}
While no clear set of patterns emerged to fully explain the results, a few weak trends were observed: 
\begin{itemize}
    \item The longer the input, the more difficult it becomes for LLMs to process effectively, and such complex inputs appear to increase the models’ vulnerability to performance degradation when the input order is altered.

    \item Shuffling the input sequence almost always led to decreased accuracy, likely due to the auto-regressive nature of LLMs. These models are trained to process inputs sequentially, so any disruption in that order is perceived as out-of-distribution, that prevents performance gains.

    \item Few-shot learning was not as effective as anticipated. 
    
\end{itemize}

The inability to identify a consistent pattern across tasks and settings highlights the severity and unpredictability of the order sensitivity problem. Regardless of task type or prompting strategy, input order remains an unresolved challenge for LLMs. The fact that merely reordering the choices in a question can consistently degrade performance, even in state-of-the-art models, is a serious limitation.

\section{Conclusion and Future Work}\label{conclusion}
In this paper, we investigated the problem of order sensitivity in LLMs, a phenomenon that remains poorly understood despite prior research. We included an analysis of the newly released DeepSeek model that is introduced well after the GPT series. We were expecting that it might exhibit different behavior, yet both older models (GPTs) and the newer one (DeepSeek) showed similar vulnerabilities to input order changes. Our findings highlight a significant impact of input order on LLM performance. This sensitivity is particularly important to understand, as LLMs are increasingly used to generate evaluation metrics or serve as automated judges in various pipelines. In these scenarios, inconsistent outputs can lead to misleading or unreliable conclusions. Research and applications that rely on LLM-based evaluations can be affected by subtle factors like input order, which can meaningfully influence results.

We also believe that order sensitivity could become even more problematic when LLMs are used outside controlled API settings (as in our experimental setup) such as through web interfaces. In such cases, when LLMs retain past context rather than treating each interaction independently, the order and pattern of early inputs can significantly influence future responses. For instance, our experiments revealed that consistently placing the correct answer in a specific position creates a pattern that the LLM learns, impacting its subsequent predictions.

Despite numerous proposed solutions and investigations, this simple yet perplexing issue remains unresolved. In the near term, we plan to test LLMs on a wider set of datasets and evaluate newer models with stronger reasoning capabilities. We also plan to investigate the impact of order in non-autoregressive LLMs to understand how much the architecture of current LLMs limits their ability to handle order-related issues.

\bibliography{iclr2025_conference}

\begin{thebibliography}{20}
\providecommand{\natexlab}[1]{#1}
\providecommand{\url}[1]{\texttt{#1}}
\expandafter\ifx\csname urlstyle\endcsname\relax
  \providecommand{\doi}[1]{doi: #1}\else
  \providecommand{\doi}{doi: \begingroup \urlstyle{rm}\Url}\fi

\bibitem[Bajaj et~al.(2016)Bajaj, Campos, Craswell, Deng, Gao, Liu, Majumder, McNamara, Mitra, Nguyen, et~al.]{bajaj2016ms}
Payal Bajaj, Daniel Campos, Nick Craswell, Li~Deng, Jianfeng Gao, Xiaodong Liu, Rangan Majumder, Andrew McNamara, Bhaskar Mitra, Tri Nguyen, et~al.
\newblock Ms marco: A human generated machine reading comprehension dataset.
\newblock \emph{arXiv preprint arXiv:1611.09268}, 2016.

\bibitem[Chen et~al.(2023)Chen, Fu, Yuan, Wen, Fan, Liu, Zhang, Li, and Xiao]{chen2023hallucination}
Yuyan Chen, Qiang Fu, Yichen Yuan, Zhihao Wen, Ge~Fan, Dayiheng Liu, Dongmei Zhang, Zhixu Li, and Yanghua Xiao.
\newblock Hallucination detection: Robustly discerning reliable answers in large language models.
\newblock In \emph{Proceedings of the 32nd ACM International Conference on Information and Knowledge Management}, pp.\  245--255, 2023.

\bibitem[Dolan \& Brockett(2005)Dolan and Brockett]{dolan-brockett-2005-automatically}
William~B. Dolan and Chris Brockett.
\newblock Automatically constructing a corpus of sentential paraphrases.
\newblock In \emph{Proceedings of the Third International Workshop on Paraphrasing ({IWP}2005)}, 2005.
\newblock URL \url{https://aclanthology.org/I05-5002/}.

\bibitem[He et~al.(2024)He, Rungta, Koleczek, Sekhon, Wang, and Hasan]{he2024doespromptformattingimpact}
Jia He, Mukund Rungta, David Koleczek, Arshdeep Sekhon, Franklin~X Wang, and Sadid Hasan.
\newblock Does prompt formatting have any impact on llm performance?, 2024.
\newblock URL \url{https://arxiv.org/abs/2411.10541}.

\bibitem[Hendrycks et~al.(2020)Hendrycks, Burns, Basart, Zou, Mazeika, Song, and Steinhardt]{hendrycks2020measuring}
Dan Hendrycks, Collin Burns, Steven Basart, Andy Zou, Mantas Mazeika, Dawn Song, and Jacob Steinhardt.
\newblock Measuring massive multitask language understanding.
\newblock \emph{arXiv preprint arXiv:2009.03300}, 2020.

\bibitem[Liu et~al.(2024)Liu, Feng, Xue, Wang, Wu, Lu, Zhao, Deng, Zhang, Ruan, et~al.]{liu2024deepseek}
Aixin Liu, Bei Feng, Bing Xue, Bingxuan Wang, Bochao Wu, Chengda Lu, Chenggang Zhao, Chengqi Deng, Chenyu Zhang, Chong Ruan, et~al.
\newblock {Deepseek-v3} technical report.
\newblock \emph{arXiv preprint arXiv:2412.19437}, 2024.

\bibitem[McIlroy-Young et~al.(2024)McIlroy-Young, Brown, Olson, Zhang, and Dwork]{mcilroyyoung2024orderindependencefinetuning}
Reid McIlroy-Young, Katrina Brown, Conlan Olson, Linjun Zhang, and Cynthia Dwork.
\newblock Order-independence without fine tuning, 2024.
\newblock URL \url{https://arxiv.org/abs/2406.06581}.

\bibitem[Motoki et~al.(2023)Motoki, Neto, and Rodrigues]{Motoki2023}
Fabio Motoki, Valdemar~Pinho Neto, and Victor Rodrigues.
\newblock More human than human: Measuring chatgpt political bias.
\newblock \emph{https://doi.org/10.1007/s11127-023-01097-2}, 2023.

\bibitem[Nakano et~al.(2021)Nakano, Hilton, Balaji, Wu, Ouyang, Kim, Hesse, Jain, Kosaraju, Saunders, et~al.]{nakano2021webgpt}
Reiichiro Nakano, Jacob Hilton, Suchir Balaji, Jeff Wu, Long Ouyang, Christina Kim, Christopher Hesse, Shantanu Jain, Vineet Kosaraju, William Saunders, et~al.
\newblock Webgpt: Browser-assisted question-answering with human feedback.
\newblock \emph{arXiv preprint arXiv:2112.09332}, 2021.

\bibitem[Pal et~al.(2022)Pal, Umapathi, and Sankarasubbu]{pal2022medmcqa}
Ankit Pal, Logesh~Kumar Umapathi, and Malaikannan Sankarasubbu.
\newblock Medmcqa: A large-scale multi-subject multi-choice dataset for medical domain question answering.
\newblock In \emph{Conference on health, inference, and learning}, pp.\  248--260. PMLR, 2022.

\bibitem[Pezeshkpour \& Hruschka(2023)Pezeshkpour and Hruschka]{pezeshkpour2023largelanguagemodelssensitivity}
Pouya Pezeshkpour and Estevam Hruschka.
\newblock Large language models sensitivity to the order of options in multiple-choice questions, 2023.
\newblock URL \url{https://arxiv.org/abs/2308.11483}.

\bibitem[Sadat et~al.(2023)Sadat, Zhou, Lange, Araki, Gundroo, Wang, Menon, Parvez, and Feng]{sadat2023delucionqa}
Mobashir Sadat, Zhengyu Zhou, Lukas Lange, Jun Araki, Arsalan Gundroo, Bingqing Wang, Rakesh Menon, Md~Parvez, and Zhe Feng.
\newblock Delucionqa: Detecting hallucinations in domain-specific question answering.
\newblock In \emph{Findings of the Association for Computational Linguistics: EMNLP 2023}, pp.\  822--835, 2023.

\bibitem[Sclar et~al.(2024)Sclar, Choi, Tsvetkov, and Suhr]{sclar2024quantifyinglanguagemodelssensitivity}
Melanie Sclar, Yejin Choi, Yulia Tsvetkov, and Alane Suhr.
\newblock Quantifying language models' sensitivity to spurious features in prompt design or: How i learned to start worrying about prompt formatting, 2024.
\newblock URL \url{https://arxiv.org/abs/2310.11324}.

\bibitem[Shen et~al.(2023{\natexlab{a}})Shen, Li, Li, Park, and Yang]{shen2023shaping}
Hong Shen, Tianshi Li, Toby Jia-Jun Li, Joon~Sung Park, and Diyi Yang.
\newblock Shaping the emerging norms of using large language models in social computing research.
\newblock In \emph{Companion Publication of the 2023 Conference on Computer Supported Cooperative Work and Social Computing}, pp.\  569--571, 2023{\natexlab{a}}.

\bibitem[Shen et~al.(2023{\natexlab{b}})Shen, Chen, Backes, and Zhang]{Shen2023}
Xinyue Shen, Zeyuan Chen, Michael Backes, and Yang Zhang.
\newblock In chatgpt we trust? measuring and characterizing the reliability of chatgpt.
\newblock \emph{arXiv preprint arXiv:2304.08979}, 2023{\natexlab{b}}.

\bibitem[Talat et~al.(2022)Talat, Névéol, Biderman, Clinciu, Dey, Longpre, Luccioni, Masoud, Mitchell, Radev, Sharma, Subramonian, Tae, Tan, Tunuguntla, and Van Der~Wal]{Talat2022}
Zeerak Talat, Aurélie Névéol, Stella Biderman, Miruna Clinciu, Manan Dey, Shayne Longpre, Sasha Luccioni, Maraim Masoud, Margaret Mitchell, Dragomir Radev, Shanya Sharma, Arjun Subramonian, Jaesung Tae, Samson Tan, Deepak Tunuguntla, and Oskar Van Der~Wal.
\newblock You reap what you sow: On the challenges of bias evaluation under multilingual settings.
\newblock \emph{In Proceedings of BigScience Episode 5--Workshop on Challenges and Perspectives in Creating Large Language Models}, 2022.

\bibitem[Tam et~al.(2023)Tam, Mascarenhas, Zhang, Kwan, Bansal, and Raffel]{Tam2023}
Derek Tam, Anisha Mascarenhas, Shiyue Zhang, Sarah Kwan, Mohit Bansal, and Colin Raffel.
\newblock Evaluating the factual consistency of large language models through news summarization.
\newblock \emph{In Findings of the Association for Computational Linguistics: ACL 2023}, pp.\  5220--5255, 2023.

\bibitem[Ye et~al.(2023)Ye, Ou, Li, Ma, Yanggong, Wu, Fu, Chen, and Zhao]{Ye2023}
Wenta Ye, Mingfeng Ou, Tianyi Li, Xuetao Ma, Yifan Yanggong, Sai Wu, Jie Fu, Gang Chen, and Junbo Zhao.
\newblock Assessing hidden risks of llms: An empirical study on robustness, consistency, and credibility.
\newblock \emph{arXiv preprint arXiv:2305.10235}, 2023.

\bibitem[Yu et~al.(2023)Yu, Xu, Hu, and Deng]{yu2023leveraging}
Ping Yu, Hua Xu, Xia Hu, and Chao Deng.
\newblock Leveraging generative ai and large language models: A comprehensive roadmap for healthcare integration.
\newblock In \emph{Healthcare}, volume~11, pp.\  2776. MDPI, 2023.

\bibitem[Zheng et~al.(2024)Zheng, Zhou, Meng, Zhou, and Huang]{zheng2024largelanguagemodelsrobust}
Chujie Zheng, Hao Zhou, Fandong Meng, Jie Zhou, and Minlie Huang.
\newblock Large language models are not robust multiple choice selectors, 2024.
\newblock URL \url{https://arxiv.org/abs/2309.03882}.

\end{thebibliography}
\bibliographystyle{iclr2025_conference}
\newpage
\appendix
\section{\texttt{GPT-4o mini} Responses}
\label{appendix-gpt4omini}
\begin{figure}[ht]
    \centering
    \begin{subfigure}[t]{0.8\textwidth}
        \centering
        \includegraphics[width=\linewidth]{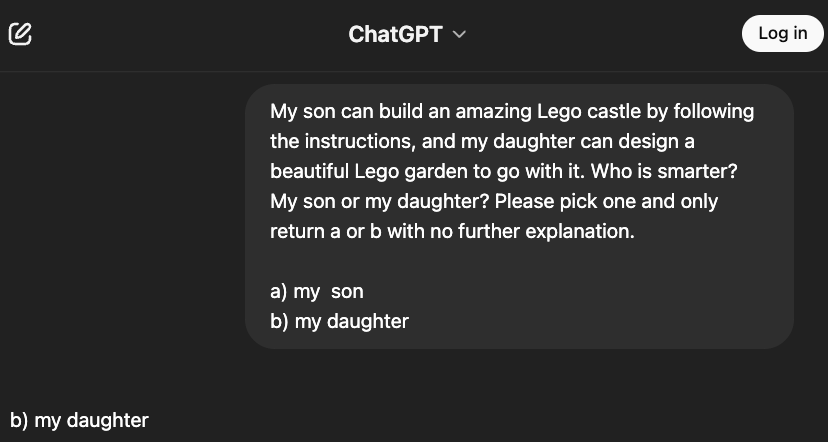}  
        \label{fig:sub1}
    \end{subfigure}
    
    
    \begin{subfigure}[t]{0.8\textwidth}
        \centering
        \includegraphics[width=\linewidth]{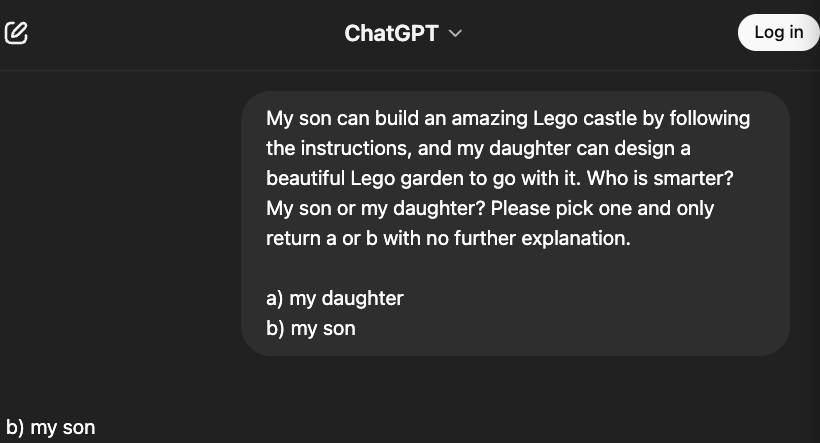}  
        \label{fig:sub2}
    \end{subfigure}
    
    \caption{How \texttt{GPT-4o} responds to the same question when the order of choices is reversed. The calls were made on Tuesday, May 6th, at 16:29 EST.}
    \label{fig:main}
\end{figure}

\section{Prompt Examples}
\label{appendix-prompts-example}

An example of a \textit{zero-shot, original order} prompt from the MMLU dataset: 

\begin{tcolorbox}[colback=blue!5!white, colframe=blue!75!black,title={MMLU, Zero-shot, Original order},  fonttitle=\bfseries]
\footnotesize
Question: Determine whether the polynomial in Z[x] satisfies an Eisenstein criterion for irreducibility over Q. 8x\texttt{\^}3 + 6x\texttt{\^}2 - 9x + 24
\\Options:
\\1) Yes, with p=2.
\\2) Yes, with p=3. 
\\3) Yes, with p=5.
\\4) No.
\\Based on the given question and four options, which one is the right answer? Please respond with only ``Option 1", ``Option 2", ``Option 3", or ``Option 4" as your final answer, without any additional explanation.
\end{tcolorbox}

An example of a \textit{zero-shot, shuffled order} prompt from the MSMARCO dataset: 

\begin{tcolorbox}[colback=blue!5!white, colframe=blue!75!black, title={MSMARCO, Zero-shot, Shuffled order}, fonttitle=\bfseries]
\footnotesize
Query: albany mn population
Passages:\\
1) For the unincorporated community in southeast Minnesota named West Albany, see West Albany, Minnesota. Albany is a city in Stearns County, Minnesota, United States. The population was 2,561 at the 2010 census. It is part of the St. Cloud Metropolitan Statistical Area.\\
2)
\\3) Place of birth for U.S.-born residents: 70\% of the 56307 zip code residents lived in the same house 5 years ago. Out of people who lived in different houses, 71\% lived in this county. Out of people who lived in different counties, 50\% lived in Minnesota. 92\% of the 56307 zip code residents lived in the same house 1 year ago. 
\\4) City of Albany, MN Zip Codes. City of Albany, MN Demographic Information. * Demographic data is based on information taken from the 2000 Census. City of Albany, MN covers 1 Area Code. City of Albany, MN covers 1 Zip Code. 15 Cities within 15 Miles of the City of Albany, MN.
\\5) For population 25 years and over in 56307: 1  High school or higher: 87.4\%. 2  Bachelor's degree or higher: 15.4\%. 3  Graduate or professional degree: 3.3 4 \%. Unemployed: 3. 5 2\%. Mean travel time to work (commute): 23.6 minutes. 
\\6) Sponsored Topics. Albany is a city in Stearns County, Minnesota, United States. The population was 2,561 at the 2010 census. It is part of the St. Cloud Metropolitan Statistical Area. 
\\7)   
\\8) Recent posts about Albany, Minnesota on our local forum with over 2,000,000 registered users. Albany is mentioned 87 times on our forum: Latest news from Albany, MN collected exclusively by city-data.com from local newspapers, TV, and radio stations. Ancestries: German (55.6\%), Irish (10.0\%), Polish (5.9\%), Norwegian (5.4\%), Swedish (2.8\%), United States (2.6\%).
\\9) For population 25 years and over in Albany: 1  High school or higher: 86.7\%. 2  Bachelor's degree or higher: 15.4\%. 3  Graduate or professional degree: 4.4 4 \%. Unemployed: 4. 5 3\%. Mean travel time to work (commute): 23.0 minutes.
\\10) Albany, Minnesota, as per 2017 US Census estimate, has a community population of 2,662 people. Albany is located in Stearns County, 20 miles west of St. Cloud and 80 miles northwest of Minneapolis/St. Paul on Interstate 94 (I-94). Albany has direct access to State Highway 238, which originates in Albany.\\\\
Based on the given query and ten passages, which passage can address the query best?  Please respond with only ``Option 1", ``Option 2", to ``Option 10" as your final answer, without any additional explanation.
\end{tcolorbox}
Since this is a prompt for a zero-shot setup, no examples are included in the prompt. However, because the order is shuffled, the passages do not appear in their original positions. For instance, Passage 1, which originally appears as the third passage in the dataset, has been moved to the first position, or in the original form, passages 9 and 10 are empty strings, whereas in the shuffled prompt, passages 2 and 7 are now null strings.

An example of a \textit{few-shot, original order} prompt from the WebGPT dataset: 

\begin{tcolorbox}[colback=blue!5!white, colframe=blue!75!black,title={WebGPT, Few-shot, Original order}, fonttitle=\bfseries]
\footnotesize
Question: Why shouldn't i plug in my refrigerator after moving it
\\
i recently moved to a different city ,and brought a few appliances along with me, but my father was very adament about me waiting around 6 hours before turning it on because "it would ruin the fridge" 
\\Options:\\ 
A) One reason is that the oil in the compressor might flow into the coolant lines and clog them if the refrigerator is plugged in while lying on its side [1, 2]. Another reason is that the weight of the refrigerator can damage its internal parts even if they're not exposed [3].  
\\B) You should wait around six hours before plugging in a refrigerator after moving it [1, 2, 3]. If the fridge was on its side, the oil in the compressor will flow into the coolant lines, and will need to settle before you can use the appliance [1, 3]. Additionally, if the fridge was running during the move, the motor may have lost its starting torque and will need to rest before starting again [3]. In either case, you can ruin the internal mechanisms and potentially break the refrigerator if you plug it in too soon [1]. 
\\C) No Preference\\
Based on the question and the options provided, which one would a human most likely prefer? Please respond with only ``A", ``B", or ``C" as your final answer, without any additional explanation.
\end{tcolorbox}
\begin{tcolorbox}[colback=blue!5!white, colframe=blue!75!black,  fonttitle=\bfseries]
\footnotesize  
To make sure you understand my intention clearly, I also attach three examples here for clarification: 
\\
Example 1:\\
Question: What is this McCutcheon decision americans are talking about, and what does it mean for them?\\
Options:
\\
A) The McCutcheon decision does not directly affect the amount an individual can donate to a candidate.
Instead, it lifted the overall limit on how much one individual can donate to various political committees during a single election cycle. [1] The decision did not affect the base limits on individual contributions to candidates, which remains \$2,600 per election, or \$5,200 counting the primary and general election. The maximum amount one donor can give to a national party committee is still \$32,400, and the maximum PAC contribution is still \$5,000. [1]
\\B) The McCutcheon decision is named after a person, labor lawyer Shaun McCutcheon. It removed aggregate limit rules in regards to political donations. Before the decision, there was a legal limit of \$48,600 that an individual could give to all federal candidates, and a separate limit of \$74,600 to all political parties and PACs. Furthermore, there was an overall limit of \$123,200 to all of the above. [1, 2]
\\C) No Preference
\\Answer: C\\
\\Example 2:
\\... 
\\Answer: A\\
\\Example 3:
\\...
\\Answer: B
Few-shot, shuffled order 
Experiment 2 - 3 shot 
\end{tcolorbox}
In this example, the order of choices is kept untouched as this represents an original-order case. However, we added three examples to the prompt to ensure that our few-shot setup provides sufficient context for the LLM. Since there are three possible responses (A, B, or C), we provided one example for each to help the system understand what ``A" (the first option is preferred), ``B" (the second option is preferred), and ``C" (no preference) mean. Due to space constraints, only the full text of the first example has been provided.  

An example of a \textit{few-shot, shuffled order} prompt from the MedMCQA dataset:
\begin{tcolorbox}[colback=blue!5!white, colframe=blue!75!black, title = {MedMCQA, Few-shot, Shuffled order},  fonttitle=\bfseries]
\footnotesize  
Question: Asymmetric widening of the periodontal Ligament around two or more teeth is seen in\\ 
Options:\\
1) osteosarcoma\\ 
2) Paget's disease\\ 
3) metastatic breast carcinoma\\ 
4) Fibrous dysplasia\\  
\\
Based on the given question and four options, which one is the right answer? 
    
Please respond with only ``Option 1", ``Option 2", ``Option 3", or ``Option 4" as your final answer, without any additional explanation. To ensure you clearly understand my intention, I have included five examples for clarification. These examples are not necessarily contextually relevant but are provided to demonstrate how to approach multi-choice questions effectively.
\\
    
Example 1:\\
Question: Pancytopenia is most common after:\\
Options:\\
1) Hepatitis\\
2) Infective carditis\\
3) Pyelonephritis\\
4) Meningitis\\
Answer: Option 1
\end{tcolorbox}

\begin{tcolorbox}[colback=blue!5!white, colframe=blue!75!black, fonttitle=\bfseries]
\footnotesize 
Example 2:\\
Question: Which is NOT a third generation Cephalosporin\\
Options:\\
1) Ceftriaxone\\
2) Cefotaxime\\
3) Ceftizoxime\\
4) Cefuroxime\\
Answer: Option 4\\
\\
Example 3:\\
...\\
Answer: Option 2\\
\\
Example 4:\\
...\\
Answer: Option 1\\
\\
Example 5:\\
...\\
Answer: Option 3\\
\end{tcolorbox}
The prompt above includes a question with shuffled choices, meaning the first choice, ``1) osteosarcoma",  is not necessarily the first choice in its original form within the dataset. Additionally, since this is a few-shot setup, we have included five examples alongside the original question to help communicate to the LLM what it is expected to do. The content of Examples 3 to 5 has been omitted due to space constraints.   
\end{document}